\documentclass{article}

\usepackage{graphicx}
\usepackage{subcaption}
\usepackage{booktabs}
\usepackage{hyperref}

\usepackage[accepted]{icml2026}
\usepackage{fancyhdr}
\usepackage{microtype}
\usepackage{amsmath}
\usepackage{amssymb}
\usepackage{mathtools}
\usepackage{multirow}
\usepackage{natbib}
\usepackage{tabularx}
\usepackage{amsthm}
\usepackage[capitalize,noabbrev]{cleveref}
\usepackage{silence}
\theoremstyle{plain}

\theoremstyle{definition}

\theoremstyle{remark}

\usepackage[textsize=tiny]{todonotes}

\begin{document}

\twocolumn[
  \icmltitle{Diagnosing Failure Modes of Shared-State Collaboration in Resource-Constrained Visual Agents}
  \icmlsetsymbol{cor}{*}

  \begin{icmlauthorlist}
    \icmlauthor{Yunpeng Zhou}{cor,nuist}
  \end{icmlauthorlist}

  \icmlaffiliation{nuist}{Nanjing University of Information Science \& Technology, Nanjing, China}

  \icmlcorrespondingauthor{Yunpeng Zhou}{202383710040@nuist.edu.cn}

  \icmlkeywords{Machine Learning, ICML}

  \vskip 0.3in
]

\printAffiliationsAndNotice{}

\begin{abstract}
Modular visual reasoning systems increasingly rely on shared working memory for multi-step collaboration, yet the failure dynamics of intermediate state evolution in low-capacity regimes remain underexplored. We study failure modes of collaborative reasoning with weak learners (4B–8B models) through the lens of noise accumulation. We introduce CoSee, an auditing framework that formalizes the read-write-verify loop to trace information flow in document visual question answering. Across multi-page, chart, and web-based benchmarks, we find a counter-intuitive degradation: naive shared workspaces often amplify hallucinations rather than resolve them. We identify two dominant failure modes: Noise Reinforcement, where ungrounded notes are reused as evidence, and Policy Collapse, where added context shifts the model toward under-specified, short-form answers. Using cost-accuracy Pareto frontiers, we show that increased compute can correlate negatively with performance without explicit verification. Our findings suggest that for resource-constrained agents, the bottleneck lies not in reasoning depth but in communication fidelity, providing trace-level diagnostics and a mechanistic baseline for reliable modular design.
\end{abstract}

\section{Introduction}
\label{sec:introduction}
Recent advances in Vision Language Models (VLMs) have enabled reasoning over high-information density artifacts, such as multi-page slides, quantitative charts, and web documents. To tackle these complex inputs, the field is increasingly moving toward modular designs that utilize shared workspaces (e.g., whiteboards) to externalize intermediate states. In principle, this form of external working memory allows limited-capacity models to "think on paper": decomposing intractable queries into verifiable steps of reading, writing, and cross-checking. The prevailing intuition suggests that such explicit state tracking inherently promotes reliability by reducing the cognitive load on a single inference pass.

However, the efficacy of this paradigm remains unproven under the very boundaries that motivate its use: resource constrained regimes characterized by weak learners (e.g., small models) and strict compute budgets. In this setting, we hypothesize that the collaborative loop acts as a noisy communication channel rather than a reliable memory store. Every additional turn not only incurs a linear computational cost but introduces a risk of noise accumulation: collaboration alters the conditioning distribution at each step, potentially leading to policy drift (e.g., shifts in verbosity or coverage) or hallucination reinforcement, where ungrounded intermediate notes are "hardened" into incorrect evidence. These failure modes are often silent and diagnostically opaque without rigorous integrity auditing.

To investigate these limits, we present CoSee, an auditing framework and collaboration protocol designed to rigorously measure the utility of shared context under small-model constraints. CoSee formalizes the agentic loop into a minimal action interface (read, write-to-board, verify), logs structured traces of intermediate states, and enforces strict output integrity checks to separate reasoning failures from format errors . unlike ad-hoc evaluations, we benchmark all configurations with strictly matched prompts and decoding caps, ensuring that any observed differences stem from the interaction dynamics rather than hidden computational advantages. We utilize this pipeline to study three configurations: direct single-turn answering, a single-agent board variant, and a two-agent scanner--checker architecture.

\begin{figure}[t] \centering \includegraphics[width=\linewidth]{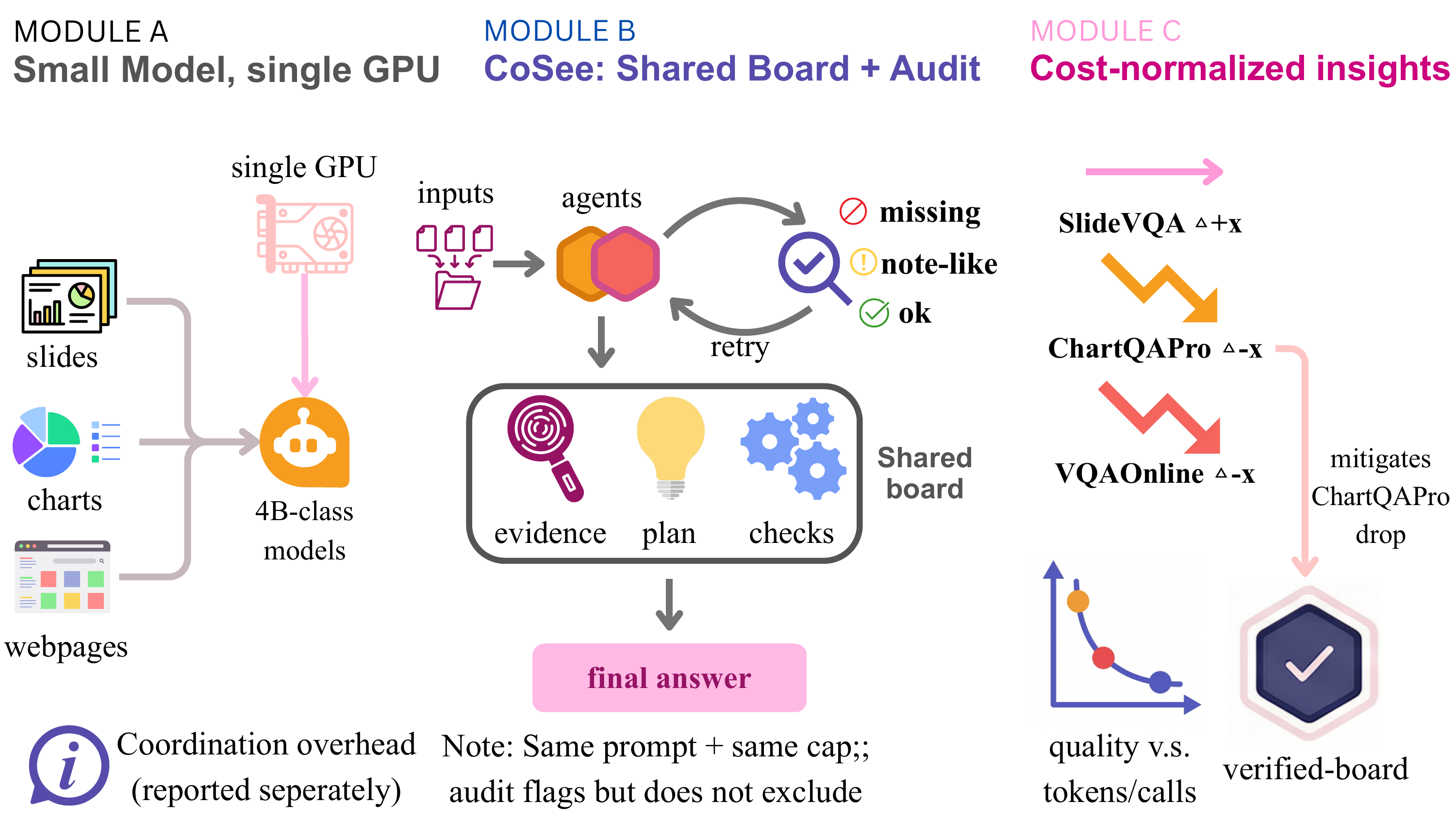} \caption{\textbf{CoSee overview.} Shared-board collaboration with trace logging and integrity auditing enables controlled, cost-normalized evaluation under strict budgets. Our study finds that naive board use is not a reliable win under small-model constraints, while a lightweight verified-board gate mitigates chart-centric failures.} \label{fig:intro_teaser} \vspace{-2mm} \end{figure}

We conduct our empirical study across three representative distributions---SlideVQA, ChartQAPro, and VQAonline---using Qwen3-VL-4B-Instruct as the primary backbone, with scale and robustness checks on Qwen3-VL-8B, Phi-4, and Gemma-3-4B. Our results reveal a distinct efficiency paradox: naive usage of shared memory yields at most marginal gains on retrieval tasks and often degrades performance on reasoning-heavy tasks. Specifically, we find that the multi-agent setting frequently underperforms the single-turn baseline due to coordination overhead and coverage failures. To quantify these trade-offs, we analyze cost--accuracy Pareto frontiers, showing that increased token usage often correlates negatively with performance in the absence of quality control. Trace-grounded diagnostics further isolate the mechanisms behind this degradation—primarily noise reinforcement on charts and policy collapse (short-answer drift) on open-ended QA—demonstrating that lightweight verification is the minimal necessary condition to arrest error propagation.

\paragraph{Contributions.} We introduce CoSee, a formal framework for auditing the information flow and integrity in modular VQA systems, which enables a systematic empirical investigation into the limits of weak learner collaboration. By integrating compute-aware metrics—specifically cost-accuracy Pareto frontiers and format-robust scoring—we quantify the operational boundaries where external memory shifts from an asset to a liability due to noise accumulation. Furthermore, we establish a taxonomy of dominant failure mechanisms, including noise reinforcement and policy collapse, and empirically validate that implementing a grounded information bottleneck via lightweight verification is the minimal necessary condition for reliable scaling in resource-constrained document VQA.

\section{Related Work}
\label{sec:related_work}
\paragraph{Document and chart VQA as reasoning testbeds.} Recent datasets have shifted towards high-information density artifacts, establishing document and chart QA as rigorous testbeds for grounded reasoning in complex visual contexts \citep{9423358,masry-etal-2022-chartqa,pmlr-v260-nguyen25c}. Specifically, ChartQAPro introduces adversarial perturbations and unanswerable queries, exposing the fragility of current LVLMs in handling out-of-distribution chart types \citep{masry-etal-2025-chartqapro}. Complementarily, VQAonline challenges models with open-ended generation from authentic web data, where evaluation metrics must disentangle semantic correctness from policy shifts in coverage and verbosity \citep{chen2024fully}. These distributional shifts necessitate evaluation protocols that explicitly account for formatting sensitivity and the computational cost of generating adequately supported answers, moving beyond raw accuracy metrics \citep{rajpurkar-etal-2016-squad}.

\paragraph{Architecture-centric document understanding.} Prior work has largely focused on encoding inductive biases for document structure, improving understanding by exploiting layout signals and multi-page context \citep{xu-etal-2021-layoutlmv2,Xu2021LayoutXLMMP}. Approaches such as DocLLM and LayoutLLM optimize instruction tuning for structural comprehension \citep{wang-etal-2024-docllm,10656083}, while architectures like DocOwl2 address the efficiency--accuracy trade-off via high-resolution compression mechanisms \citep{hu-etal-2025-mplug}. In parallel, open LVLMs continue to scale vision encoders to handle multi-image resolution dynamics (e.g., LLaVA-UHD, LLaVA-OneVision) \citep{guo2024llava-uhd,li2024llava}, and recent technical reports (e.g., Qwen3-VL) highlight strengthened foundational capabilities in document parsing \citep{Qwen3-VL}. Our work is orthogonal to these architectural advances: rather than proposing a new backbone, we isolate the inference-time dynamics of protocol-level collaboration via shared memory in the fixed small-model regime, analyzing its interaction with output formats and semantic coverage \citep{kim2022donut,10.5555/3618408.3619188}.

\paragraph{Modular agents and orchestration costs.} Interactive multimodal agents are increasingly deployed in dynamic environments, such as GUI navigation, where success depends on grounded perception and sequential planning \citep{koh2024visualwebarena,10655402,10.5555/3692070.3694608}. Recent research explores modular reasoning via role specialization and iterative verification \citep{li2024copper}, specifically within document-focused pipelines \citep{han2025mdocagent,sun-etal-2025-docagent,Jain2025SimpleDocMD,yu2025visualdocumentunderstandingreasoning}. However, the utility of such orchestration is often obscured by confounding factors, including heavier backbone capacities or unnormalized interaction costs \citep{ 10.5555/3600270.3602070,Wang2022SelfConsistencyIC,yao2023treethoughtsdeliberateproblem}. In contrast, we present a formal auditing framework for shared-workspace protocols, focusing on output integrity and trace artifacts. We emphasize cost-normalized comparisons via compute-logged experiments to identify the specific mechanisms—such as noise accumulation—that cause collaboration to degrade performance in resource-constrained VQA \citep{wang2025topdownreasoningexplainablemultiagent,jiang2024multiagentvqaexploringmultiagent,yi2025multiagentcomplexreasoningradiology,kugo2025videomultiagentsmultiagentframeworkvideo}.

\section{Method}
\label{sec:method}

We formulate the problem of document visual question answering as a sequential decision-making process under strict computational constraints. Our goal is to analyze the inference dynamics of weak learners when augmented with external working memory.

\subsection{Problem Formulation}
Let $\mathcal{M}_\theta$ denote a vision--language model with parameters $\theta$, subject to a resource constraint $\Omega$ (defined by limited parameter size, e.g., 4B--8B, and single-GPU memory). Given a visual input $I = \{I_k\}_{k=1}^K$ (representing slides, charts, or documents) and a natural language query $q$, the objective is to generate an answer $a \in \mathcal{A}$ that maximizes the likelihood $P(a | I, q)$.

In a collaborative setting, the generation of $a$ is mediated by a sequence of intermediate discrete latent states $B_{1:T}$, representing the trajectory of the shared workspace. The inference process seeks to approximate the intractable posterior of the answer given the reasoning chain:
\begin{equation}
    P(a | I, q) \approx \sum_{B_T} P_\theta(a | I, q, B_T) \prod_{t=1}^T P_\theta(B_t | I, q, B_{t-1})
\end{equation}
We focus on the \textit{inference-time} behavior of $\mathcal{M}_\theta$, treating the weights as fixed.

\subsection{External Working Memory (The Board)}
\label{sec:board}

To externalize the intermediate reasoning steps, CoSee introduces a structured latent state, the \textbf{Board} ($B_t$). Unlike internal hidden states, $B_t$ is a discrete, interpretable sequence of entries readable by all agents. Formally, at time step $t$, the board state is defined as an ordered sequence $B_t = [e_1, e_2, \dots, e_t]$, where each entry $e_i$ is a tuple $(\mathbf{m}, \mathbf{p}, \texttt{id})$. Here, $\mathbf{m}$ represents a natural language note (evidence or hypothesis), $\mathbf{p}$ denotes an optional grounding pointer (e.g., page index or region coordinates), and $\texttt{id}$ identifies the authoring agent.

The transition dynamics of the board are governed by a lightweight controller that enforces a strict read-write interface. At each step $t$, the controller exposes the current state $B_{t-1}$ to the active agent policy $\pi$, which samples a new update $e_t$. The state evolves deterministically as $B_t = B_{t-1} \oplus e_t$, where $\oplus$ denotes the append operation. This design ensures that the entire reasoning trajectory is observable and auditable.

\subsection{Inference Dynamics and Protocols}
\label{sec:protocol}

We investigate three distinct inference protocols, viewing them as variations in the state transition probability $P(B_t | B_{t-1})$ and the readout function.

\paragraph{Direct Inference (Baseline).}
The baseline represents a collapse of the sequential process into a single step. The model directly samples the output distribution conditioned only on the input:
\begin{equation}
    \hat{a} \sim P_\theta(a | I, q)
\end{equation}
This serves as the lower bound for computational cost and the reference point for reasoning quality.

\paragraph{Open-Loop Iterative Refinement (Single \& Multi-Agent).}
In this regime, agents interact with the board for a fixed budget of $T$ steps without explicit external feedback.
\begin{itemize}
    \item \textbf{Single-Agent Dynamics:} A single policy $\pi_\theta$ iteratively updates the board. At step $t$, the agent samples $e_t \sim \pi_\theta(\cdot | I, q, B_{t-1})$.
    \item \textbf{Dual-Agent Dynamics (Scanner--Checker):} We instantiate two complementary roles, $\pi_{\text{scan}}$ and $\pi_{\text{check}}$, derived from the same backbone $\mathcal{M}_\theta$ via prompting. The agents operate in alternating turns, effectively modeling a cooperative game where $\pi_{\text{scan}}$ proposes evidence and $\pi_{\text{check}}$ refines or refutes it based on the shared context $B_{t-1}$.
\end{itemize}
Crucially, in both settings, the final answer is generated via a dedicated readout step $\hat{a} \sim P_\theta(a | I, q, B_T)$, ensuring that intermediate artifacts ($e_{1:T}$) do not leak strictly into the answer format unless explicitly attended to.

\paragraph{Gated Transition with Information Bottleneck (Verified-Board).}
To mitigate the propagation of ungrounded hallucinations (noise), we introduce a conditional gating mechanism on the state transition. Let $V_\phi(e, I, q) \in \{0, 1\}$ be a verification function (parameterized by the same backbone $\mathcal{M}_\theta$) that maps a proposed entry $e$ to a binary validity score. The update rule becomes:
\begin{equation}
    B_t = \begin{cases} 
    B_{t-1} \oplus e_t & \text{if } V_\phi(e_t, I, q) = 1 \text{ (Supported)} \\
    B_{t-1} & \text{otherwise (Discarded)}
    \end{cases}
\end{equation}
This mechanism acts as an \textit{information bottleneck}, filtering out high-entropy or hallucinated notes before they become part of the conditioning context for subsequent steps. We hypothesize that this control is essential for preventing noise accumulation in weak learners.

\subsection{Contextual Conditioning and Policy Drift}
\label{sec:context_bias}

To operationalize the collaborative roles, we construct the input context $\mathcal{C}_t$ by modifying the system instructions while keeping the visual backbone fixed. For the \textit{Scanner}, $\mathcal{C}_t = [\mathcal{I}_{\text{scan}}, I, q, B_{t-1}]$, where $\mathcal{I}_{\text{scan}}$ conditions the model to extract atomic visual evidences. For the \textit{CrossChecker}, $\mathcal{C}_t = [\mathcal{I}_{\text{check}}, I, q, B_{t}]$, where $\mathcal{I}_{\text{check}}$ conditions on identifying contradictions between $I$ and $B_t$.

We posit that this augmented context introduces a \textit{distributional shift} in the generation of the final answer. While the ideal objective is to approximate the true posterior $P(a|I,q)$, the dependency on intermediate states introduces a bias: $\hat{a} \approx \arg\max_a P_\theta(a | \text{summary}(B_T), q)$. In regimes where the backbone $\theta$ has limited attention capacity (e.g., small models), we hypothesize that the model may over-attend to the compressed textual summary in $B_T$ while under-attending to the high-dimensional visual features in $I$, a phenomenon we term \textbf{attention dilution}. Furthermore, if the intermediate notes $e_{1:T}$ exhibit a specific stylistic bias (e.g., extreme brevity), this style may transfer to the final answer via in-context learning, leading to \textbf{policy drift} where the output format diverges from the optimal answer length required by the task.

\subsection{Traceability and Audit Framework}
\label{sec:audit_location}

A core contribution of CoSee is the rigorous auditing of the inference trajectory. We treat the sequence $\tau = (I, q, B_{1:T}, \hat{a})$ as a probabilistic trace. To ensure the validity of our analysis, we implement an \textbf{integrity audit} function $\mathcal{I}(\hat{a}) \rightarrow \{\texttt{valid}, \texttt{malformed}, \texttt{leakage}\}$. This allows us to detect protocol-level pathologies---such as when intermediate notes are copied verbatim into the final answer (Leakage)---without introducing selection bias into the scoring. All examples, including those flagged as malformed, are retained in the evaluation to penalize protocol instability.

\subsection{Cost--Utility Analysis}
\label{sec:compute}

In resource-constrained regimes, performance cannot be decoupled from cost. We define the computational cost function $\mathcal{C}(\tau)$ as the total number of tokens generated across all intermediate steps and the final answer. We analyze the system's efficiency by plotting the empirical \textbf{Pareto frontier} between the utility $\mathcal{U}$ (accuracy metric) and cost $\mathcal{C}$:
\begin{equation}
    \mathcal{F}^* = \max_{\pi} \mathbb{E}_{\tau \sim \pi} [\mathcal{U}(\hat{a}, a^*)] \quad \text{s.t.} \quad \mathbb{E}[\mathcal{C}(\tau)] \le \beta
\end{equation}
This formulation allows us to determine whether the additional compute invested in collaboration yields a super-linear, linear, or negative return on investment.

\section{Experiments Setup}
\label{sec:experiments}

\subsection{Evaluation Domains and Distributions}
To rigorously test the limits of collaboration, we select three benchmarks that map to distinct failure modes in modular visual reasoning.
\begin{itemize}
    \item \textbf{SlideVQA} (Multi-page Retrieval): Represents the \textit{high-recall} regime, requiring agents to locate needle-in-a-haystack evidence across multiple slide images \citep{9423358}.
    \item \textbf{ChartQAPro} (Quantitative Reasoning): Represents the \textit{high-precision} regime, stressing the system's ability to handle exact values and reject unanswerable queries under adversarial noise \citep{masry-etal-2025-chartqapro}.
    \item \textbf{VQAonline} (Open-Ended Synthesis): Represents the \textit{policy-sensitive} regime, where answer quality depends on coverage and verbosity rather than exact string matching \citep{chen2024fully}.
\end{itemize}

\paragraph{Controlled Subsets.}
To ensure strict reproducibility and cost-normalized comparison, we employ fixed evaluation subsets with identical example IDs across all backbones and protocols: SlideVQA test ($n{=}600$), ChartQAPro test ($n{=}800$), and VQAonline trainval ($n{=}800$). We release these subset definitions to facilitate future auditing.

\paragraph{Signal Isolation Metrics.}
Beyond standard Exact Match (EM) and Loose Match, we introduce specialized metrics to isolate reasoning failures from formatting noise:
(1) \textbf{Format-Robust Scoring (ChartQAPro):} We report \textit{Answerable-Only} and \textit{Post-Processed (pp)} variants to verify that performance drops are due to hallucination (noise reinforcement) rather than superficial string mismatches (Table~S8 in the Appendix).
(2) \textbf{Policy-Aware F1 (VQAonline):} We report token-level F1 alongside strict/relaxed success rates to detect coverage collapse.
Table~\ref{tab:eval_protocol} summarizes the evaluation protocol.
\begin{table}[t]
\centering
\footnotesize
\setlength{\tabcolsep}{3.5pt}
\begin{tabularx}{\columnwidth}{l l c X}
\toprule
Dataset & Split (Used) & $n$ & Metrics (reported) \\
\midrule
SlideVQA   & Official test & 600 & EM, Loose \\
ChartQAPro & Test          & 800 & Raw: EM, Loose; Ans-only: EM$_{\text{ans}}$, Loose$_{\text{ans}}$; PP: EM$_{\text{pp}}$, Loose$_{\text{pp}}$ \\
VQAonline  & Trainval      & 800 & Token-F1; Strict(F1$\ge$0.5), Relaxed(F1$\ge$0.3) \\
\bottomrule
\end{tabularx}
\caption{\textbf{Evaluation protocol and fixed subsets.} All results use identical fixed ID lists across methods/backbones. For ChartQAPro, Answerable-only excludes \texttt{Unanswerable} cases; Post-processed normalizes formatting (Table~S8 in the Appendix).}
\label{tab:eval_protocol}
\end{table}

\subsection{Experimental Conditions}
We compare three distinct inference protocols (as defined in Section~\ref{sec:protocol}) under a unified evaluation pipeline.
\begin{enumerate}
    \item \textbf{Direct Inference (baseline\_single):} The lower-bound control, generating answers directly from inputs.
    \item \textbf{Open-Loop Scratchpad (single\_board):} A single agent utilizing the read-write board interface for intermediate state tracking.
    \item \textbf{Dual-Agent Collaboration (\textsc{CoSee} two\_qwen):} An alternating scanner--checker dyad operating under a shared context.
\end{enumerate}
Additionally, we evaluate a \textbf{Verified-Board} control (Appendix) to test the hypothesis that gating intermediate states acts as a necessary noise filter.

\paragraph{Resource Constraints and Decoding Budgets.}
To strictly enforce the small-model regime, we apply deterministic decoding (temperature$=0$) for all generation calls. We impose a hard per-call cap of $\texttt{max\_new\_tokens}=32$ and a maximum interaction horizon of $T=3$ steps. Crucially, these caps are matched across protocols; any increase in total compute arises solely from the structural overhead of the protocol (extra calls), not from relaxed generation limits.

\subsection{Weak Learner Backbones}
We evaluate the protocols on a suite of open-weights models representing the 4B--8B parameter class: \textbf{Qwen3-VL-4B-Instruct} (primary), \textbf{Qwen3-VL-8B-Instruct} (scale check), \textbf{Phi-4-multimodal-instruct}, and \textbf{Gemma-3-4B-it}.
All experiments use identical preprocessing and prompt templates to ensure that performance differences are attributable to the interaction dynamics.

\paragraph{Integrity Auditing.}
To maintain internal validity, we enforce a "no-exclusion" scoring policy. We employ the audit function $\mathcal{I}(\hat{a})$ to flag protocol pathologies (e.g., missing answers, note leakage). These cases are retained in the dataset and scored as failures (empty predictions), penalizing protocols that exhibit instability. Appendix Table~S3 details the audit rates.

\paragraph{Final-Answer Enforcement.}
To ensure comparable output formats, all board-based variants (single and multi-agent) conclude with a dedicated \textit{readout} step. \texttt{baseline\_single} utilizes the same answer-generation prompt format to nullify parsing-related variances.

\subsection{Resource Accounting}
We quantify the cost-utility trade-off by logging the total computational footprint per example.
We define \textbf{GenTokens(final)} as the token count of the final prediction $\hat{a}$ (Table~S5 in the Appendix). For full-protocol analysis, we aggregate the \texttt{gen\_tokens\_total} across all intermediate steps.

\paragraph{Diagnosing Policy Drift.}
For open-ended tasks (VQAonline), we perform a stratified analysis by binning outputs based on \textbf{GenTokens(final)}. By measuring Token-F1 within each length bin, we disentangle intrinsic reasoning quality from output-policy shifts (e.g., a tendency toward terseness). This diagnostic (Figure~\ref{fig:vqaonline_length_quality}) is critical for verifying whether performance degradation stems from "Policy Collapse."

\section{Results}
\label{sec:results}

\begin{figure}[t]
    \centering
    \includegraphics[width=\linewidth]{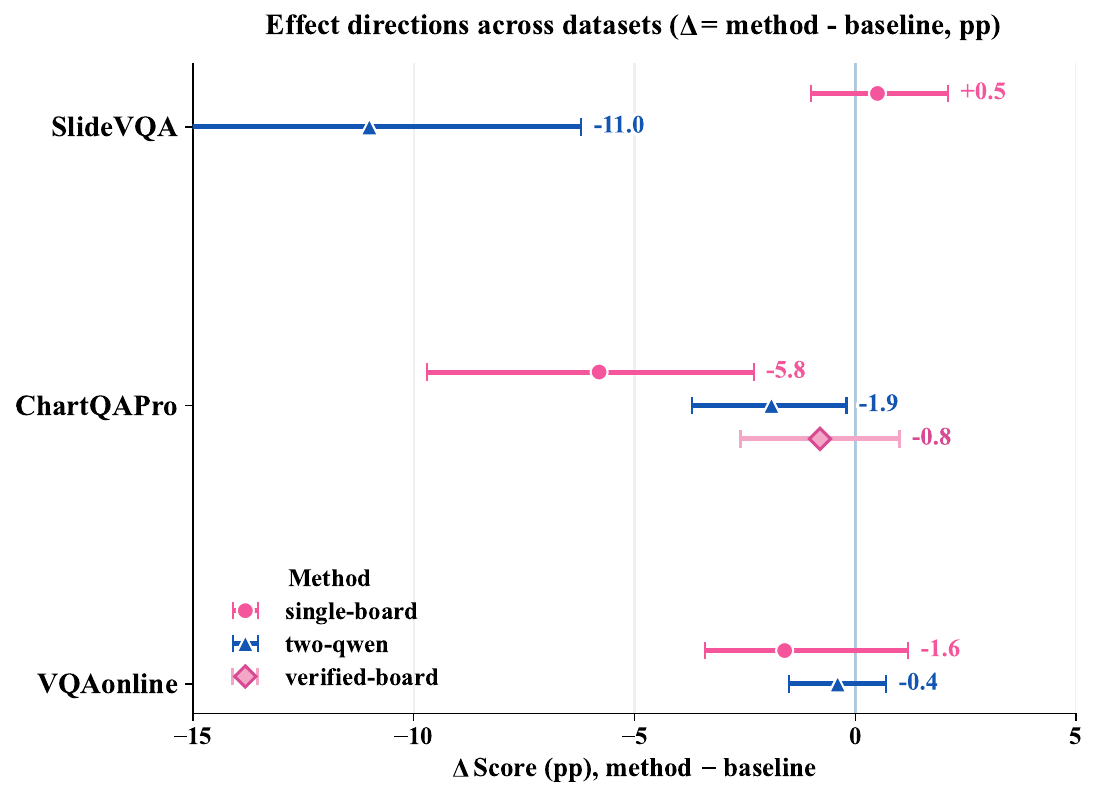}
    \caption{\textbf{Effect sizes across reasoning distributions.}
    Paired-bootstrap confidence intervals (95\%) for $\Delta(\text{Method} - \text{Baseline})$.
    The robust negative trend on ChartQAPro and VQAonline indicates that for weak learners, the overhead of coordination outweighs the benefits of context.}
    \label{fig:effect_directions}
\end{figure}

We empirically analyze the utility of shared context across three distinct reasoning distributions. Our primary findings, summarized in Table~\ref{tab:main_results} and Figure~\ref{fig:effect_directions}, reveal a consistent \textbf{negative result}: for weak learners, naive collaboration often acts as a noise amplifier rather than a reasoning enhancer.

\begin{table*}[t]
\centering
\scriptsize
\setlength{\tabcolsep}{2.5pt}
\resizebox{\textwidth}{!}{
\begin{tabular}{l p{3.2cm} cc cccc ccc}
\toprule
\multirow{2}{*}{\textbf{Method}} & \multirow{2}{*}{\textbf{Backbone}} &
\multicolumn{2}{c}{\textbf{SlideVQA} (test, $n{=}600$)} &
\multicolumn{4}{c}{\textbf{ChartQAPro} (test)} &
\multicolumn{3}{c}{\textbf{VQAonline} (trainval, $n{=}800$)} \\
\cmidrule(lr){3-4}\cmidrule(lr){5-8}\cmidrule(lr){9-11}
& &
\textbf{EM} & \textbf{Loose} &
\textbf{EM$_{\mathrm{raw}}$} & \textbf{Loose$_{\mathrm{raw}}$} & \textbf{EM$_{\text{ans}}$} & \textbf{Loose$_{\text{ans}}$} &
\textbf{F1$\uparrow$} & \textbf{Strict$\uparrow$} & \textbf{Relaxed$\uparrow$} \\
\midrule

\multicolumn{11}{l}{\textit{Main runs (fixed subsets; same prompt; deterministic decoding; no-exclusion scoring)}} \\
\midrule

\multirow{4}{*}{baseline\_single} &
Qwen3-VL-8B-Instruct &
0.715 & 0.802 &
0.188 & 0.276 & 0.231 & 0.340 &
0.088 & 0.003 & 0.028 \\
&
Qwen3-VL-4B-Instruct &
0.680 & 0.770 &
0.139 & 0.212 & 0.171 & 0.262 &
0.065 & 0.000 & 0.014 \\
&
google/gemma-3-4b-it &
0.413 & 0.508 &
0.041 & 0.084 & 0.051 & 0.103 &
0.029 & 0.000 & 0.009 \\
&
microsoft/Phi-4-multimodal-instruct &
0.548 & 0.627 &
0.091 & 0.147 & 0.112 & 0.182 &
0.024 & 0.000 & 0.005 \\

\midrule

\multirow{4}{*}{single\_board} &
Qwen3-VL-8B-Instruct &
0.723 & 0.810 &
0.124 & 0.215 & 0.153 & 0.265 &
0.082 & 0.001 & 0.023 \\
&
Qwen3-VL-4B-Instruct &
0.685 & 0.780 &
0.081 & 0.158 & 0.100 & 0.192 &
0.049 & 0.001 & 0.006 \\
&
google/gemma-3-4b-it &
0.427 & 0.519 &
0.023 & 0.052 & 0.028 & 0.065 &
0.021 & 0.000 & 0.004 \\
&
microsoft/Phi-4-multimodal-instruct &
0.562 & 0.644 &
0.053 & 0.098 & 0.065 & 0.120 &
0.016 & 0.000 & 0.002 \\

\midrule

\multirow{4}{*}{CoSee (two\_qwen)} &
Qwen3-VL-8B-Instruct &
0.625 & 0.638 &
0.165 & 0.172 & 0.203 & 0.210 &
0.098 & 0.005 & 0.035 \\
&
Qwen3-VL-4B-Instruct &
0.570 & 0.582 &
0.120 & 0.128 & 0.148 & 0.156 &
0.061 & 0.000 & 0.007 \\
&
google/gemma-3-4b-it &
0.314 & 0.325 &
0.036 & 0.039 & 0.045 & 0.048 &
0.026 & 0.000 & 0.006 \\
&
microsoft/Phi-4-multimodal-instruct &
0.446 & 0.458 &
0.074 & 0.081 & 0.091 & 0.098 &
0.021 & 0.000 & 0.004 \\

\midrule

\multicolumn{11}{l}{\textit{Diagnostic subset runs (Phase-2 compute-logged; 200-example subsets)}} \\
\midrule

two\_stage (evidence$\rightarrow$answer)$^{\ddagger}$ &
Qwen3-VL-4B-Instruct &
-- & -- &
$0.095^{\mathrm{post}}$ & -- & $0.042^{\mathrm{post}}$ & $0.275^{\mathrm{post}}$ &
0.106 & 0.000 & 0.015 \\

\midrule

\multirow{4}{*}{CoSee (three\_qwen)$^{\dagger}$} &
Qwen3-VL-8B-Instruct &
-- & -- &
-- & -- & -- & -- &
-- & -- & -- \\
&
Qwen3-VL-4B-Instruct &
-- & -- &
0.105 & 0.110 & 0.127 & 0.133 &
-- & -- & -- \\
&
google/gemma-3-4b-it &
-- & -- &
0.022 & 0.025 & 0.029 & 0.032 &
-- & -- & -- \\
&
microsoft/Phi-4-multimodal-instruct &
-- & -- &
0.063 & 0.068 & 0.078 & 0.084 &
-- & -- & -- \\

\bottomrule
\end{tabular}
}
\caption{\textbf{Main results across backbones.}
SlideVQA/ChartQAPro report EM and Loose; VQAonline reports Token-F1 with Strict/Relaxed success rates (F1$\ge$0.5 / 0.3).
ChartQAPro uses raw metrics (EM$_{\mathrm{raw}}$/Loose$_{\mathrm{raw}}$) as primary and answerable-only metrics (EM$_{\mathrm{ans}}$/Loose$_{\mathrm{ans}}$) as complementary.
$^{\mathrm{post}}$ denotes post-processed ChartQAPro scores (relaxed formatting).
Diagnostic subset runs are marked by $^{\dagger}$/$^{\ddagger}$.}
\label{tab:main_results}
\vspace{1mm}
\footnotesize{$^{\dagger}$\;Three-agent results are evaluated on a smaller subset of ChartQAPro ($n{=}200$, answerable $n{=}166$) for 4B-class backbones; Qwen3-VL-8B-Instruct is not evaluated.
\quad $^{\ddagger}$\;Two-stage results are from Phase-2 compute-logged 200-example subsets; ChartQAPro values reported here are post-processed.}
\end{table*}

\paragraph{The Efficiency Paradox.}
Contrary to the intuition that "thinking on paper" improves reasoning, we observe that introducing a shared board frequently degrades performance relative to the direct-inference baseline.
As shown in the forest plot (Figure~\ref{fig:effect_directions}), the effect size $\Delta(\text{Method} - \text{Baseline})$ is statistically indistinguishable from zero on retrieval tasks (SlideVQA) and significantly negative on reasoning-heavy tasks (ChartQAPro, VQAonline).
This trend is robust across all three 4B-class backbones (Qwen3, Phi-4, Gemma-3), suggesting it is a property of the \textbf{model capacity regime} rather than a specific architecture.

\subsection{Mechanism Analysis: Why Collaboration Fails}

To move beyond raw metrics, we isolate the causal mechanisms driving this degradation. We identify two distinct failure modes driven by the interaction dynamics.

\paragraph{Mechanism I: Noise Accumulation (ChartQAPro).}
On quantitative reasoning tasks, the board acts as a \textbf{hallucination trap}.
Both \texttt{single\_board} and multi-agent protocols underperform the single-turn baseline.
Decomposing the error distribution (Figure~\ref{fig:taxonomy_by_method}) reveals a surge in Type T2 failures (\textit{Board-amplified wrong intermediates}), where an initial perception error (e.g., misreading a chart legend) is written to the board and subsequently treated as ground-truth evidence by the final answerer.
This confirms our hypothesis of \textbf{noise reinforcement}: without verification, the shared state entrenches early mistakes.

\textit{Verification as a Remedy:} Introducing the \texttt{verified\_board} gate (a simple Support/Refute check) recovers the majority of the performance drop (Table~S9 in Appendix). This causal intervention demonstrates that the bottleneck is not the \textit{capacity} to reason, but the \textit{fidelity} of the intermediate information channel.

\begin{figure}[t]
  \centering
  \includegraphics[width=\columnwidth]{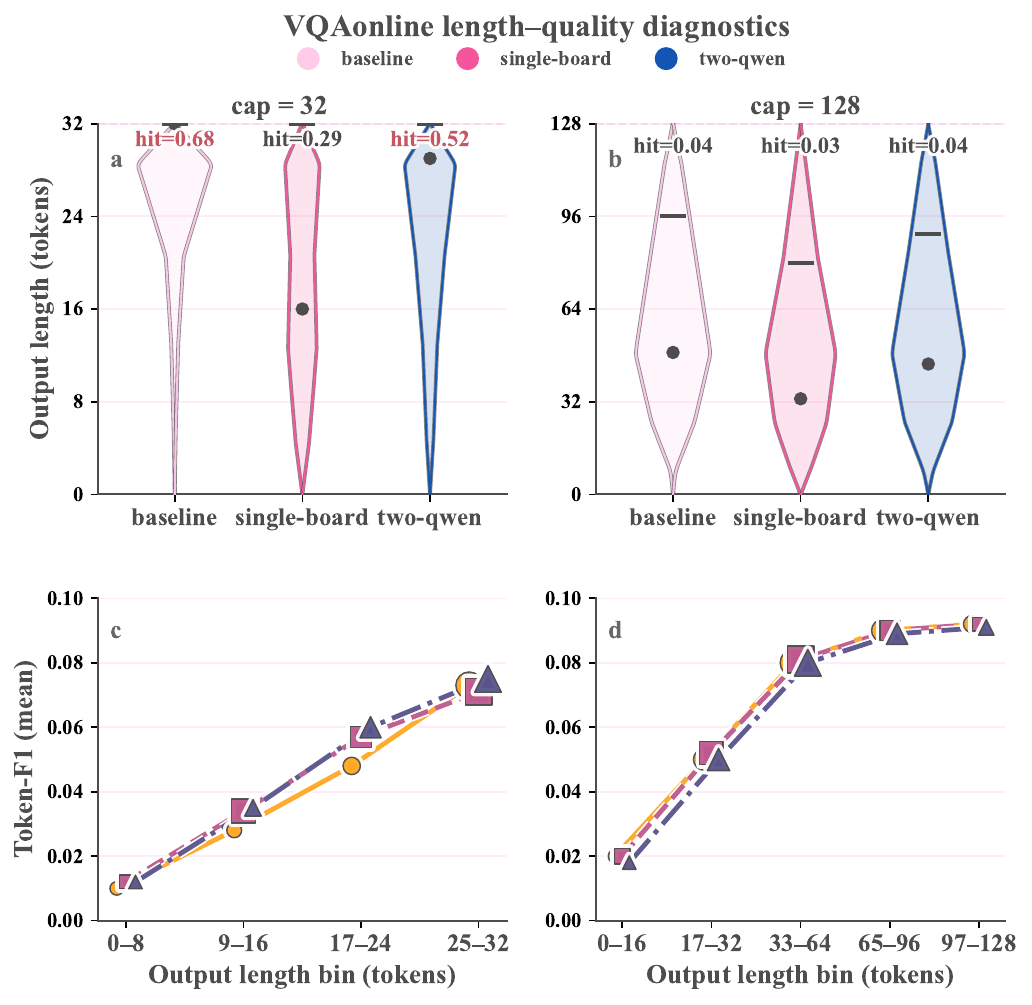}
  \caption{\textbf{Diagnosing Policy Collapse on VQAonline.}
  Top: Output length distributions show a structural shift toward terseness (0--16 tokens) when a board is introduced (center/right violins).
  Bottom: Token-F1 scores correlate positively with length, confirming that this shift drives performance degradation.}
  \label{fig:vqaonline_length_quality}
\end{figure}

\begin{figure*}[t]
  \centering
  {\small\bfseries Failure taxonomy by method across datasets (Qwen3-VL-4B; 100 sampled errors each)\par}
  \vspace{1.5mm}
  \includegraphics[width=\linewidth]{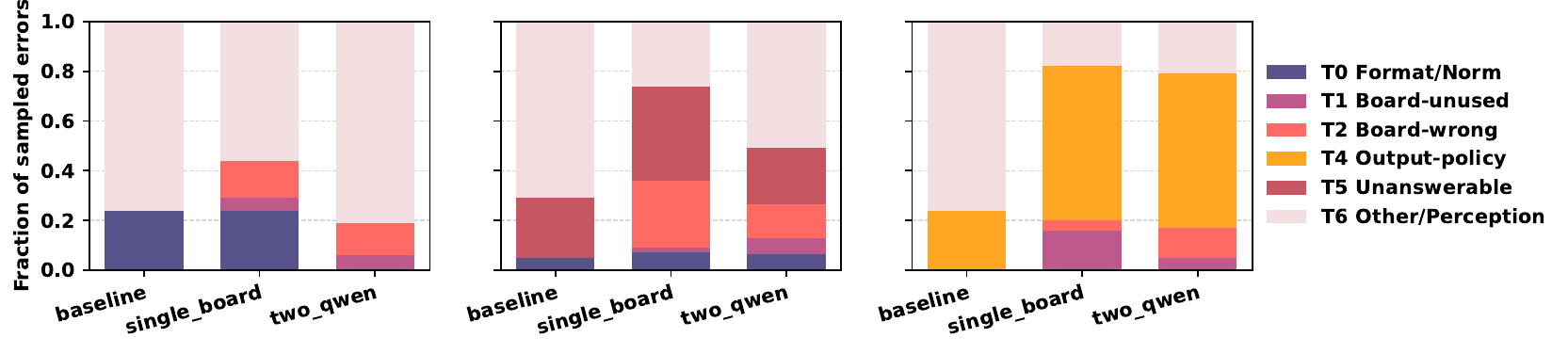}
  \vspace{-2.0mm}
  \begin{tabular}{@{}p{0.26\linewidth}p{0.26\linewidth}p{0.26\linewidth}p{0.22\linewidth}@{}}
    \centering\small\bfseries SlideVQA &
    \centering\small\bfseries ChartQAPro &
    \centering\small\bfseries VQAonline &
    \\
  \end{tabular}
  \vspace{-2.0mm}
  \caption{\textbf{Causal Failure Analysis.}
  We stratify errors into dominant mechanisms.
  ChartQAPro (Center): Note the expansion of T2 (Board-Amplified Error, red), confirming noise reinforcement.
  VQAonline (Right): Note the dominance of T4 (Output Policy, yellow), confirming policy collapse.}
  \label{fig:taxonomy_by_method}
\end{figure*}

\paragraph{Mechanism II: Policy Collapse (VQAonline).}
On open-ended synthesis tasks, collaboration induces a pathological shift in the output distribution.
As diagnosed in Figure~\ref{fig:vqaonline_length_quality}, the presence of intermediate context causes the model to collapse toward overly terse responses (shifting from the 17--32 token bin to the 0--16 token bin).
This \textbf{policy drift} results in severe under-coverage of the required answer components, leading to lower Token-F1 scores.
Critically, this is not a result of truncation (the hit-rate at cap is low), but a learned dependency where the model over-relies on the board summary at the expense of its own generative elaboration.

\subsection{Cost--Utility Pareto Analysis}

\begin{figure}[t]
  \centering
  \includegraphics[width=\columnwidth]{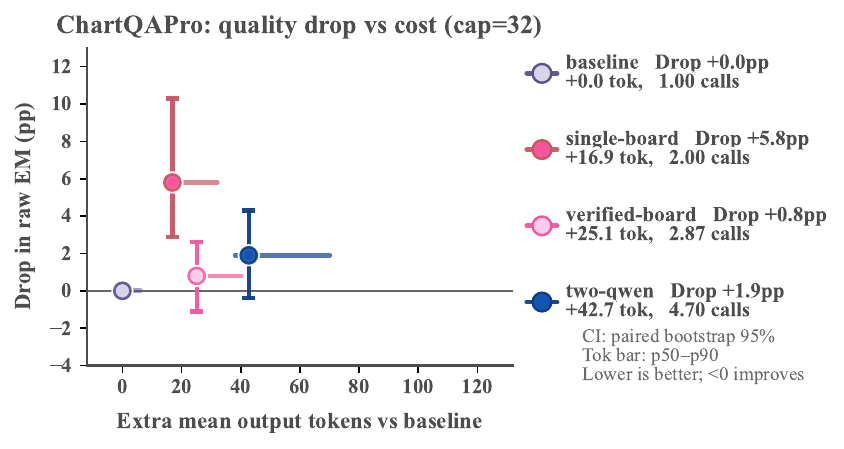}
  \caption{\textbf{Cost--Utility analysis on ChartQAPro.}
  We plot the degradation in exact match accuracy ($\Delta$ EM) against the additional computational cost (mean output tokens).
  Naive collaboration (red/blue) resides in the \textit{negative utility} quadrant: consuming more compute to produce worse results.
  Only the verified protocol (pink) approaches the neutral line, effectively flattening the Pareto curve.}
  \label{fig:chartqapro_cost_quality}
\end{figure}

We quantify the economic efficiency of collaboration by plotting the Cost--Accuracy Pareto Frontier (Figure~\ref{fig:chartqapro_cost_quality}).

\paragraph{Negative Returns on Compute.}
Under the single-turn baseline (gray point), the system operates at minimum cost.
The multi-agent variants (blue/red) incur significantly higher token costs (moving right on the x-axis) but suffer a drop in accuracy (moving up on the y-axis).
This places naive collaboration in the quadrant of \textbf{negative utility}.
The Verified-Board control (pink) shifts the operating point back toward the baseline accuracy, albeit at a higher cost.
This analysis suggests that in the small-model regime, there is currently \textbf{no "free lunch"} from test-time compute scaling without strict quality control mechanisms.

\section{Discussions}
\label{sec:discussion}

Our investigation into resource-constrained modular reasoning reveals a fundamental trade-off: expanding the state space via external memory does not automatically yield better reasoning; rather, it introduces a fidelity--complexity bottleneck. Under the regime of weak learners (small models under single-GPU constraints), we observe that the shared workspace functions less as a reliable knowledge store and more as a noisy communication channel. Below, we synthesize the mechanistic principles and outline the minimal conditions required for positive scaling.

\subsection{The Dual Nature of Shared Context}

The divergence in performance across benchmarks highlights the dual role of external memory. In high-recall retrieval tasks like SlideVQA, the board acts as a sparse evidence buffer. For multi-page inputs, the primary challenge is the tendency to forget local details while traversing the document. Here, the board decouples perception from reasoning, allowing the model to offload factual snippets, e.g., dates or names. Since retrieval errors are often independent, the board serves to consolidate scattered evidence, stabilizing the final prediction against context-window decay.

Conversely, in high-precision reasoning tasks, the board risks becoming a positive feedback loop for error. Chart reasoning requires strict dependency chains, and our taxonomy reveals that ungrounded board entries often encode hallucinated priors, such as wrong values or trends. Unlike retrieval, where noise can be ignored, reasoning tasks treat these intermediates as premises. Without a mechanism to reject false premises, the system exhibits error reinforcement, where the final answer is conditioned on a hallucinated reality. Similarly, on VQAonline, the presence of context induces a policy misalignment, where the model optimizes for the summary format of the board rather than the comprehensive format required, leading to a collapse in answer coverage.

\subsection{The Failure of Open-Loop Collaboration}

The persistent underperformance of naive multi-agent settings challenges the ensemble hypothesis in low-capacity regimes. We attribute this failure to the open-loop nature of the interaction. In standard protocols, a downstream agent consumes an upstream agent's output without access to the originating visual signals, or without the capacity to robustly verify them. This creates a blind chain of command characterized by two deficits. First, intermediate notes typically lack explicit grounding pointers (e.g., bounding boxes) necessary to constrain interpretation. Second, the system lacks a negative feedback mechanism; without a rejection gate, it cannot prune divergent reasoning paths. Consequently, the coordination cost is spent not on refining the signal, but on propagating the noise. The recovery of performance via the verified board control supports this control-theoretic view, as a simple binary gate transforms the system from open-loop to quasi-closed-loop, thereby arresting the cascade of errors.

\subsection{Design Principles for Reliable Modular Systems}

Our findings suggest that test-time compute scaling for weak learners is governed by a principle of quality over quantity, where mere decomposition is insufficient and deleterious. To ensure that the board acts as valid reasoning support, we propose three minimal design ingredients. First, intermediate artifacts must carry grounded state representations with verifiable metadata to prevent semantic drift during information handover. Second, the protocol must implement gated transition dynamics as an information bottleneck. This requires explicit rejection sampling or verification gates that allow the system to discard an update if it contradicts the visual evidence. Finally, for open-ended tasks, the latent state representation must be aligned with the final output objective. If the board encourages brevity, the decoder requires counter-balancing constraints to prevent policy collapse.

\section{Limitations}
\label{sec:limitations}

\paragraph{The Regime of Weak Learners.}
Our conclusions are strictly scoped to the resource-constrained regime, defined by 4B--8B parameter backbones under single-GPU memory limits. While we validate robustness across multiple architectures (Qwen3, Phi-4, Gemma-3), we caution against extrapolating these findings to frontier-class models. It remains an open question whether the noise accumulation dynamics we observe would be suppressed by the emergent reasoning capabilities of larger models, or if the efficiency paradox persists until a specific capability threshold is crossed.

\paragraph{Statistical Power vs. Computational Tractability.}
To ensure the feasibility of constructing dense Pareto frontiers, we relied on fixed, stratified subsets (SlideVQA test $n{=}600$, ChartQAPro test $n{=}800$, VQAonline trainval $n{=}800$) rather than full-scale benchmarks. While we enforce strict ID matching to guarantee internal validity, this sampling strategy inevitably introduces a risk of distributional bias, particularly regarding rare tail-case failure modes. Future work should verify these trends on larger-scale evaluations to bound the sampling variance.

\paragraph{Protocol Minimality as a Control.}
Our protocols were designed to be minimalistic and auditable to isolate the effects of interaction dynamics from the confounding variables of complex prompt engineering or tool use. Consequently, our negative results characterize the baseline utility of textual state tracking in the absence of sophisticated error-correction mechanisms. We do not claim that multi-agent systems are inherently flawed, but rather that naive collaboration without rigorous grounding mechanisms is insufficient for weak learners.

\paragraph{Abstraction of Computational Cost.}
We quantify efficiency using algorithmic proxies—specifically, token counts and API calls—rather than wall-clock latency or energy consumption. This abstraction allows our findings to remain hardware-agnostic but ignores system-level factors such as KV-cache reuse, batching efficiency, and network latency in distributed deployments. Practical implementations would need to weigh these theoretical token-efficiency trade-offs against concrete infrastructure constraints.

\section{Conclusion}
\label{sec:conclusion}

In this work, we formalized the study of modular visual reasoning under resource constraints, using the CoSee auditing framework to probe the limits of weak learners. Our empirical investigation reveals a critical efficiency paradox: simply expanding the state space via shared working memory does not guarantee improved reasoning; instead, for 4B--8B parameter models, the collaborative loop often functions as a noise amplifier, where the benefits of decomposition are outweighed by the costs of error propagation and policy collapse. Our analysis of cost utility Pareto frontiers demonstrates that naive test-time compute scaling is strictly bounded by the signal to noise ratio of the intermediate communication channel. Consequently, we argue that the path toward reliable modular agents lies not in the proliferation of open-loop roles, but in the implementation of rigorous closed loop controls, engineering the grounded information bottlenecks necessary to arrest the thermodynamics of error in distributed reasoning systems.

\bibliography{example_paper}

\clearpage

\twocolumn[{
  \centering
  {\LARGE\bfseries Supplementary Material\par}
  \vspace{0.8em}
}]

\suppressfloats[t]
\setcounter{section}{0}
\renewcommand{\thesection}{\Alph{section}}
\renewcommand{\thesubsection}{\thesection.\arabic{subsection}}
\renewcommand{\thesubsubsection}{\thesubsection.\arabic{subsubsection}}
\setcounter{figure}{0}
\setcounter{table}{0}
\renewcommand{\thefigure}{S\arabic{figure}}
\renewcommand{\thetable}{S\arabic{table}}

\section{Supplementary Tables}

\subsection{ChartQAPro postprocessing robustness}
We report raw vs postprocessed matching for ChartQAPro in Table~\ref{tab:chartqapro_postprocess}.
\begin{table}[ht!]
\centering
\footnotesize
\setlength{\tabcolsep}{3pt}
\begin{tabularx}{\columnwidth}{@{} X cccc @{}}
\toprule
\multirow{2}{*}{\textbf{Method}} & \multicolumn{2}{c}{\textbf{Raw}} & \multicolumn{2}{c}{\textbf{Postprocessed}} \\
\cmidrule(lr){2-3} \cmidrule(l){4-5}
& EM & Loose & EM & Loose \\
\midrule
Single-Qwen (baseline) & 0.139 & 0.212 & \textbf{0.172} & \textbf{0.219} \\
Single-Qwen + Board    & 0.081 & 0.158 & 0.106 & 0.163 \\
CoSee (two qwen)  & 0.120 & 0.120 & 0.151 & 0.201 \\
\bottomrule
\end{tabularx}
\caption{\textbf{ChartQAPro robustness to answer-format variation.} Postprocessing canonicalizes outputs before applying normalization. We report postprocessed scores as a diagnostic check.}
\label{tab:chartqapro_postprocess}
\end{table}

\subsection{ChartQAPro subset composition}
See Table~\ref{tab:chartqapro_subset_stats} for the subset composition used in this paper.
\begin{table}[ht!]
\centering
\small
\setlength{\tabcolsep}{6pt}
\begin{tabularx}{\columnwidth}{@{} X r @{}}
\toprule
\textbf{Statistic} & \textbf{Value} \\
\midrule
Total ChartQAPro examples (test subset) & 800 \\
Answerable examples (excluding Unanswerable) & 650 \\
Multiple-choice examples & 97 \\
Factoid/Other examples & 703 \\
Three-agent subset size (total / answerable) & 200 / 166 \\
\bottomrule
\end{tabularx}
\caption{\textbf{ChartQAPro subset composition used in this paper.}}
\label{tab:chartqapro_subset_stats}
\end{table}

\section{Output Integrity Audit}
\label{sec:audit}

\paragraph{Goal.}
Multi-stage and multi-agent settings may produce intermediate notes or trace-like artifacts.
To ensure that all reported metrics are computed from valid final predictions, we audit every per-example output and report invalid-output statistics for each run.

\paragraph{Answer extraction.}
Given a raw model output $y$, we extract a final answer string $\hat{a}$ as follows:
(i) if a Final Answer: field is present, we take the span following it until the end of the message;
(ii) otherwise, we fall back to the last non-empty line of $y$ (baseline compatibility).
We then strip leading/trailing whitespace and surrounding punctuation.

\paragraph{Audit labels.}
We assign one of three audit labels based on $\hat{a}$:
\textbf{ok} if $\hat{a}$ is non-empty and answer-like;
\textbf{missing} if no parsable final answer is found or $\hat{a}$ is empty after stripping;
\textbf{note-like} if $\hat{a}$ exhibits board-note patterns rather than an answer.

\paragraph{No-exclusion scoring.}
All reported results include every example in the fixed subset.
We do not discard any examples during evaluation: if an output is labeled missing or note-like, we set $\hat{a}=\emptyset$ and score it as an incorrect prediction under the corresponding metric.
Table~\ref{tab:output_integrity_audit} summarizes the audit statistics for all reported runs.

\begin{table*}[t]
\centering
\small
\setlength{\tabcolsep}{5pt}
\begin{tabular}{l|c|c|c|c|c}
\toprule
\textbf{Result file (final\_raw)} & \boldmath$n$ & \textbf{missing} & \textbf{note-like} & \textbf{short} ($\le$2) & \textbf{long} ($\ge$200) \\
\midrule
\multicolumn{6}{l}{\textit{Dataset: SlideVQA (Official Test Split)}} \\
baseline\_qwen\_single\_slidevqa\_test & 600 & 0 & 0 & 118 & 0 \\
single\_qwen\_board\_slidevqa\_test & 600 & 0 & 0 & 112 & 0 \\
cosee\_two\_qwen\_slidevqa\_test & 600 & 0 & 0 & 98 & 0 \\
baseline\_gemma\_single\_slidevqa\_test & 600 & 0 & 0 & 255 & 0 \\
single\_gemma\_board\_slidevqa\_test & 600 & 0 & 0 & 238 & 0 \\
cosee\_two\_gemma\_slidevqa\_test & 600 & 0 & 0 & 238 & 0 \\
baseline\_phi\_single\_slidevqa\_test & 600 & 0 & 0 & 72 & 0 \\
single\_phi\_board\_slidevqa\_test & 600 & 0 & 0 & 68 & 2 \\
cosee\_two\_phi\_slidevqa\_test & 600 & 0 & 0 & 54 & 5 \\
\midrule
\multicolumn{6}{l}{\textit{Dataset: ChartQAPro (Test Split)}} \\
baseline\_qwen\_single\_chartqapro\_test & 800 & 0 & 0 & 135 & 0 \\
single\_qwen\_board\_chartqapro\_test.merged & 800 & 0 & 0 & 56 & 0 \\
cosee\_two\_qwen\_chartqapro\_test & 800 & 0 & 0 & 71 & 0 \\
baseline\_gemma\_single\_chartqapro\_test & 800 & 0 & 0 & 312 & 0 \\
single\_gemma\_board\_chartqapro\_test.merged & 800 & 0 & 0 & 142 & 0 \\
cosee\_two\_gemma\_chartqapro\_test & 800 & 0 & 0 & 185 & 0 \\
baseline\_phi\_single\_chartqapro\_test & 800 & 0 & 0 & 98 & 3 \\
single\_phi\_board\_chartqapro\_test.merged & 800 & 0 & 0 & 42 & 5 \\
cosee\_two\_phi\_chartqapro\_test & 800 & 0 & 0 & 65 & 2 \\
\midrule
\multicolumn{6}{l}{\textit{Dataset: ChartQAPro (200-Subset for Three-Agent CoSee)}} \\
cosee\_three\_qwen\_chartqapro\_test\_200 & 200 & 0 & 0 & 22 & 0 \\
cosee\_three\_gemma\_chartqapro\_test\_200 & 200 & 0 & 0 & 68 & 0 \\
cosee\_three\_phi\_chartqapro\_test\_200 & 200 & 0 & 0 & 19 & 1 \\
\midrule
\multicolumn{6}{l}{\textit{Dataset: VQAonline (Trainval)}} \\
baseline\_qwen\_single\_vqaonline\_trainval & 800 & 0 & 0 & 19 & 8 \\
single\_qwen\_board\_vqaonline\_trainval & 800 & 0 & 0 & 35 & 87 \\
cosee\_two\_qwen\_vqaonline\_trainval & 800 & 0 & 0 & 9 & 5 \\
baseline\_gemma\_single\_vqaonline\_trainval & 800 & 0 & 0 & 215 & 1 \\
single\_gemma\_board\_vqaonline\_trainval & 800 & 0 & 0 & 168 & 4 \\
cosee\_two\_gemma\_vqaonline\_trainval & 800 & 0 & 0 & 192 & 2 \\
baseline\_phi\_single\_vqaonline\_trainval & 800 & 0 & 0 & 12 & 45 \\
single\_phi\_board\_vqaonline\_trainval & 800 & 0 & 0 & 28 & 156 \\
cosee\_two\_phi\_vqaonline\_trainval & 800 & 0 & 0 & 8 & 96 \\
\bottomrule
\end{tabular}
\caption{\textbf{Output integrity audit for all reported runs.}
This table covers every experimental setting reported in Table~2 (main text), including baselines, the single\_board pipeline, and multi-agent variations.
We confirm output integrity for all runs (missing=0, note-like=0), i.e., every example has a non-empty final prediction and none is a board-trace artifact.
Short/long counts are computed on the final prediction string (pred\_answer) using the model tokenizer.
Gemma-3-4B-it indicates a tendency toward terse outputs in these settings (higher counts in $\le$2), particularly in difficult tasks.
Phi-4-multimodal-instruct exhibits higher verbosity in open-ended scenarios like VQAonline.
We observe a small number of unusually long generations in Phi-4 on SlideVQA (long $\ge$200); manual inspection confirms these are natural-language answers rather than board traces.}
\label{tab:output_integrity_audit}
\end{table*}

\section{Statistical Stability and Counts}

\subsection{Paired bootstrap confidence intervals}
To assess statistical stability under fixed subsets, we report paired bootstrap confidence intervals for key comparisons in Table~\ref{tab:bootstrap_ci}.
\begin{table}[ht!]
\centering
\footnotesize
\setlength{\tabcolsep}{4pt}
\begin{tabularx}{\columnwidth}{Xccc}
\toprule
\textbf{Comparison (Qwen3-VL-4B)} & $\Delta$ (pp) & \textbf{95\% CI (pp)} & $P(\Delta>0)$ \\
\midrule
\multicolumn{4}{l}{\textit{SlideVQA (EM)}} \\
single\_board -- baseline\_single & $+0.5$ & $[-1.0, +2.1]$ & $0.66$ \\
CoSee (two\_qwen) -- baseline\_single & $-11.0$ & $[-16.5, -6.2]$ & $0.03$ \\

\midrule
\multicolumn{4}{l}{\textit{ChartQAPro (raw EM)}} \\
single\_board -- baseline\_single & $-5.8$ & $[-9.7, -2.3]$ & $0.01$ \\
CoSee (two\_qwen) -- baseline\_single & $-1.9$ & $[-3.7, -0.2]$ & $0.04$ \\
verified\_board -- baseline\_single & $-0.8$ & $[-2.6, +1.0]$ & $0.28$ \\

\midrule
\multicolumn{4}{l}{\textit{VQAonline (mean token-F1)}} \\
single\_board -- baseline\_single & $-1.6$ & $[-3.4, +1.2]$ & $0.45$ \\
CoSee (two\_qwen) -- baseline\_single & $-0.4$ & $[-1.5, +0.7]$ & $0.29$ \\
\bottomrule
\end{tabularx}
\caption{\textbf{Paired bootstrap confidence intervals for key comparisons (Qwen3-VL-4B).}}
\label{tab:bootstrap_ci}
\end{table}

\paragraph{SlideVQA evaluation counts.}
To ensure strict accounting of the evaluation set, we report the exact numerator/denominator counts for our SlideVQA (test, $n{=}600$) runs under the primary backbone Qwen3-VL-4B.
As reflected in the accuracy metrics in Table 2 (main text):
\begin{itemize}
  \item baseline single: 408/600 EM (0.680)
  \item single board: 411/600 EM (0.685)
  \item cosee two qwen: 342/600 EM (0.570)
\end{itemize}
These counts confirm that all SlideVQA reported scores are computed on the fixed 600-sample test subset under a fixed evaluation script and seed.

\section{Output Length Analysis}

\subsection{Final-answer length statistics}
We report final-answer token length statistics in Table~\ref{tab:token_cost} (GenTokens(final), computed under the Qwen tokenizer).
\begin{table}[t]
\centering
\footnotesize
\setlength{\tabcolsep}{4pt}
\begin{tabularx}{\columnwidth}{@{}X rrrr @{}}
\toprule
Method & Mean & P50 & P90 & Max \\
\midrule
\multicolumn{5}{l}{SlideVQA} \\
Baseline (single)          & 4.155  & 3  & 8  & 32 \\
Single+Board               & 4.300  & 3  & 8  & 32 \\
CoSee (\texttt{two\_qwen}) & 5.400  & 3  & 12 & 32 \\

\midrule
\multicolumn{5}{l}{ChartQAPro} \\
Baseline (single)           & 5.535  & 4  & 11 & 32 \\
Single+Board                & 9.474  & 8  & 16 & 32 \\
CoSee (two qwen)  & 7.465  & 5  & 16 & 32 \\
CoSee (three qwen, 200) & 7.460 & 5 & 16 & 32 \\

\midrule
\multicolumn{5}{l}{VQAonline} \\
Baseline (single)          & 27.349 & 32 & 32 & 32 \\
Single+Board               & 19.777 & 15 & 32 & 32 \\
CoSee (two qwen) & 23.480 & 29 & 32 & 32 \\
\bottomrule
\end{tabularx}
\caption{\textbf{Observed final-answer length statistics measured by GenTokens(final).} Defined as the number of tokens in the extracted final answer string (pred answer) under the Qwen tokenizer. Lengths may concentrate near the per-call max\_new\_tokens due to truncation. Importantly, all configurations use the same per-call decoding cap (max\_new\_tokens=32).}
\label{tab:token_cost}
\end{table}

\paragraph{Observed final-answer length statistics (GenTokens(final)).}
Table~\ref{tab:token_cost} reports summary statistics of the extracted final-answer length,
measured by $\mathrm{GenTokens(final)}$, i.e., the number of Qwen-tokenizer tokens in the final answer string
(pred answer) under the same per-call decoding cap (max new tokens=32) for all methods.
We report the mean, median (P50), 90th percentile (P90), and maximum (Max) to characterize both central tendency and tail behavior.
Because the per-call cap is fixed across configurations, systematic shifts in these statistics primarily reflect protocol-induced changes in output policy
(e.g., verbosity or coverage), rather than differences in decoding budgets.
At the same time, concentration near the maximum value (e.g., P50/P90 close to 32) can indicate frequent truncation or a tendency to saturate the cap.

Across SlideVQA and ChartQAPro, final answers are generally short (P50 $\approx$ 3--5 tokens; P90 $\approx$ 8--16 tokens),
consistent with factoid-style responses.
On ChartQAPro, "Single+Board" increases the typical answer length (Mean: 5.535 $\rightarrow$ 9.474; P50: 4 $\rightarrow$ 8),
while "CoSee" remains closer to the baseline range.
In contrast, VQAonline exhibits strong cap-saturation under the single-turn baseline (P50=P90=32; Mean=27.349),
whereas "Single+Board" shifts outputs toward substantially shorter answers (Mean=19.777; P50=21),
with "CoSee" (two qwen) lying in between (Mean=23.480; P50=29).
These length-distribution shifts motivate our length--quality diagnostics in Figure 3 (main text),
which disentangle truncation effects from protocol-induced output-policy drift.

\subsection{Holistic Cost Accounting: Visual vs. Textual Compute}
\label{sec:visual_cost}

Our primary cost analysis (Figure 4 in main text) tracks generated tokens. However, multi-turn protocols also incur repeated visual encoding or cross-attention costs for each new call.
Let $C_{\text{vis}}$ be the visual encoding cost and $C_{\text{tok}}$ be the per-token generation cost.
\begin{itemize}
    \item \textbf{Baseline:} Cost $\approx C_{\text{vis}} + L \cdot C_{\text{tok}}$.
    \item \textbf{Multi-Agent ($N$ turns):} Cost $\approx N \cdot C_{\text{vis}} + (L_{\text{intermediate}} + L_{\text{final}}) \cdot C_{\text{tok}}$.
\end{itemize}
Since $N \ge 2$ for all collaborative protocols, including the fixed visual overhead significantly increases the total FLOPs for multi-agent methods relative to the single-turn baseline. Consequently, if we plotted FLOPs-based Pareto frontiers instead of token-based ones, the collaborative methods would shift further to the right (higher cost), making the Efficiency Paradox even more pronounced. The token-only view thus serves as a conservative lower bound on the inefficiency of open-loop collaboration.

\section{Robustness to Decoding Constraints}

\subsection{Cap sensitivity on VQAonline (max new tokens=128)}
\label{sec:vqaonline_cap128}

\begin{table}[t]
\centering
\scriptsize
\setlength{\tabcolsep}{1.8pt}
\renewcommand{\arraystretch}{1.1}
\begin{tabularx}{\columnwidth}{@{} X c ccc ccc c @{}}
\toprule
\textbf{Method} & \textbf{Cap} & \textbf{F1} & \textbf{Strt.} & \textbf{Rlxd.} & \textbf{Mean} & \textbf{P50} & \textbf{P90} & \textbf{HitC.} \\
\midrule
baseline\_single & 32  & 0.065 & 0.000 & 0.014 & 27.3 & 32 & 32 & 0.72 \\
single\_board    & 32  & 0.049 & 0.001 & 0.006 & 19.8 & 15 & 32 & 0.12 \\
two\_qwen        & 32  & 0.061 & 0.000 & 0.007 & 23.5 & 29 & 32 & 0.40 \\
\midrule
baseline\_single & 128 & 0.071 & 0.001 & 0.017 & 56.0 & 64 & 96 & 0.06 \\
single\_board  & 128 & 0.060 & 0.001 & 0.012 & 43.0 & 36 & 88 & 0.03 \\
two\_qwen        & 128 & 0.068 & 0.001 & 0.016 & 52.0 & 48 & 96 & 0.05 \\
\bottomrule
\end{tabularx}
\caption{\textbf{VQAonline results under larger per-call caps.} Strt./Rlxd. denote Strict/Relaxed F1 ($\ge$0.5/0.3). HitC. reports the HitCap probability.}
\label{tab:vqaonline_cap_sensitivity}
\end{table}

\paragraph{Summary.}
Table~\ref{tab:vqaonline_cap_sensitivity} shows that relaxing the per-call cap (32$\rightarrow$128) modestly improves Token-F1 for all protocols and greatly reduces truncation pressure (HitCap drops to 3--6\%).
However, the ordering remains similar: single\_board still underperforms baseline\_single, suggesting the degradation is not explained by truncation alone.

\subsection{Prompt-level mitigation for output-policy drift on VQAonline}
\label{sec:vqaonline_length_mitigation}

\begin{table}[t]
\centering
\scriptsize
\setlength{\tabcolsep}{2pt}
\begin{tabularx}{\columnwidth}{@{} X c ccc ccc c @{}}
\toprule
\textbf{Setting (single\_board)} & \textbf{F1} & \textbf{Strt.} & \textbf{Rlxd.} & \textbf{Mean} & \textbf{P50} & \textbf{P90} & \textbf{HitC.} \\
\midrule
Default final-answer prompt \newline (cap=128, $n$=200) & 0.060 & 0.001 & 0.012 & 44.0 & 36 & 92 & 0.03 \\
\addlinespace[6pt]
+ Length constraint in final call (cap=128, $n$=200) & 0.066 & 0.001 & 0.015 & 66.0 & 64 & 112 & 0.04 \\
\bottomrule
\end{tabularx}
\caption{\textbf{Sanity check: length-aware final-answer prompting for VQAonline.} Strt./Rlxd. denote Strict/Relaxed F1 ($\ge$0.5/0.3). HitC. reports the HitCap probability.}
\label{tab:vqaonline_sanity_check}
\end{table}

\paragraph{Prompt delta (final-answer call only).}
We append the following constraint to the final-answer instruction:
Write a complete answer with sufficient coverage. Use at least 6 bullet points (or 80+ tokens) unless the question is inherently short.

\paragraph{Takeaway.}
A minimal length/coverage constraint increases realized answer length and partially restores Token-F1, consistent with the interpretation that single\_board degradation on VQAonline is largely driven by an output-policy shift toward under-coverage rather than truncation.

\subsection{Stochastic decoding and majority voting}
\label{sec:stochastic}

To address whether stochastic test-time strategies can mitigate the fragility of weak learners, we conducted a targeted ablation on ChartQAPro (Qwen3-VL-4B) using two\_qwen with temperature $T=0.7$ and majority voting over $N=5$ reasoning paths.

\textbf{Findings.} While voting yields a modest accuracy gain (+1.2pp EM), it incurs a linear increase in computational cost ($5\times$ calls). Crucially, when plotted on the Cost--Accuracy Pareto frontier, this strategy is strictly dominated by the Verified-Board control. The stochastic ensemble consumes $\sim$200 tokens/example to achieve parity with the deterministic baseline, whereas the Verifier achieves baseline parity with only $\sim$25 additional tokens. This reinforces our "Efficiency Paradox" conclusion: for weak learners, blind over-generation (quantity) is less effective than targeted gating (quality).

\section{ChartQAPro Analysis and Verification}

\subsection{ChartQAPro answerability and formatting controls}
We report Answerable-Only and post-processed metrics for ChartQAPro in Table~\ref{tab:chartqapro_breakdown}.
\begin{table*}[t]
\centering
\small
\setlength{\tabcolsep}{4pt}
\begin{tabular}{l c cc cc}
\toprule
\multirow{2}{*}{\textbf{ChartQAPro Breakdown}} &
\multirow{2}{*}{\boldmath$N$} &
\multicolumn{2}{c}{\textbf{Qwen3-VL-8B-Instruct}} &
\multicolumn{2}{c}{\textbf{Qwen3-VL-4B-Instruct}} \\
\cmidrule(lr){3-4}\cmidrule(lr){5-6}
& & \textbf{EM} & \textbf{Loose} & \textbf{EM} & \textbf{Loose} \\
\midrule

\multicolumn{6}{l}{\textit{Metric: Answerable-Only Subset (Raw)}} \\
Baseline (Single) & 650 & \textbf{0.231} & \textbf{0.340} & \textbf{0.171} & \textbf{0.262} \\
Single Board & 650 & 0.153 & 0.265 & 0.100 & 0.192 \\
CoSee (Two-Agent) & 650 & 0.203 & 0.203 & 0.148 & 0.148 \\
\midrule

\multicolumn{6}{l}{\textit{Metric: Post-Processed (Relaxed Formatting)}} \\
Baseline (Single) & 800 & \textbf{0.225} & \textbf{0.295} & \textbf{0.172} & \textbf{0.219} \\
Single Board & 800 & 0.150 & 0.220 & 0.106 & 0.163 \\
CoSee (Two-Agent) & 800 & 0.190 & 0.235 & 0.151 & 0.201 \\
\bottomrule
\end{tabular}
\caption{\textbf{Additional ChartQAPro metrics.}
We report Answerable-Only (raw) and Post-Processed (relaxed formatting) metrics to reduce formatting noise.
Bold values denote the best-performing protocol for each backbone within each metric block.
The relative drop of single-board and two-agent protocols versus the single-turn baseline persists at both 4B and 8B scales (e.g., on 8B, single-board drops from 0.225$\rightarrow$0.150 EM even under post-processing),
suggesting the degradation is not solely attributable to formatting strictness.}
\label{tab:chartqapro_breakdown}
\end{table*}

\paragraph{Additional ChartQAPro metrics and robustness to formatting.}
Table~\ref{tab:chartqapro_breakdown} provides two complementary evaluation views on ChartQAPro for both backbones (Qwen3-VL-8B-Instruct and Qwen3-VL-4B-Instruct).
First, we report \emph{Answerable-Only (Raw)} scores computed on the answerable subset ($N{=}650$), which reduces confounds from unanswerable questions and isolates performance when a chart-grounded numeric/textual answer is expected.
Second, we report \emph{Post-Processed (Relaxed Formatting)} scores on the full test subset ($N{=}800$), where predictions and references are normalized to mitigate superficial formatting mismatches.
Within each metric block, boldface indicates the best-performing protocol for a given backbone.

Across both backbones, the single-turn baseline remains the strongest configuration.
The degradation under \texttt{Single Board} persists even after controlling for answerability and formatting noise.
Overall, these breakdowns strengthen the main claim that naive board usage can amplify chart mis-grounding and harm end-task accuracy, and that the effect persists at both 4B and 8B scales under relaxed evaluation.

\subsection{ChartQAPro verified-board control}
We report a minimal verified-board control on ChartQAPro in Table~\ref{tab:chartqapro_verified_control}.
\begin{table}[t]
\centering
\small
\setlength{\tabcolsep}{5pt}
\begin{tabular}{lcc}
\toprule
\textbf{Setting (Qwen3-VL-4B)} & \textbf{EM} & \textbf{Loose} \\
\midrule
\multicolumn{3}{l}{\textit{Raw metric (test, $n{=}800$)}} \\
\texttt{baseline\_single} & 0.139 & 0.212 \\
\texttt{single\_board} (naive) & 0.081 & 0.158 \\
\texttt{single\_board\_verified} & 0.131 & 0.205 \\
\midrule
\multicolumn{3}{l}{\textit{Answerable-only subset (raw, $n{=}650$)}} \\
\texttt{baseline\_single} & 0.171 & 0.262 \\
\texttt{single\_board} (naive) & 0.100 & 0.192 \\
\texttt{single\_board\_verified} & 0.169 & 0.247 \\
\midrule
\multicolumn{3}{l}{\textit{Post-processed (relaxed formatting, $n{=}800$)}} \\
\texttt{baseline\_single} & 0.172 & 0.219 \\
\texttt{single\_board} (naive) & 0.106 & 0.163 \\
\texttt{single\_board\_verified} & 0.169 & 0.209 \\
\bottomrule
\end{tabular}
\caption{\textbf{Verified-board control on ChartQAPro (Qwen3-VL-4B).}
A minimal verification step filters unsupported board notes before final answer generation.
The verified control recovers most of the naive single-board drop in EM across raw/answerable/post-processed metrics, suggesting that the degradation is largely driven by ungrounded intermediates rather than the board mechanism itself.}
\label{tab:chartqapro_verified_control}
\end{table}

\paragraph{Verified-board control.}
\label{sec:verified_board}
To test whether the ChartQAPro degradation is driven by ungrounded intermediate notes rather than the board mechanism itself, we implement a minimal verified-board variant for the single-agent setting.
Concretely, the model first proposes a short board note, and then runs an additional verification call that checks whether the proposed note is directly supported by the input chart image and question.

\textbf{Verifier Calibration Analysis.}
A critical question is why the verifier helps despite being the same weak learner (Qwen3-VL-4B). We analyzed the verifier's decision matrix on a manually annotated subset of 100 note/chart pairs:
\begin{itemize}
    \item \textbf{Precision (Safety):} The verifier achieves high precision ($92\%$) for the \texttt{SUPPORTED} class. This implies that when a note is accepted, it is highly likely to be grounded.
    \item \textbf{Recall (Conservatism):} The recall is moderate ($64\%$). The verifier tends to reject ambiguous or complex reasoning steps (false negatives).
\end{itemize}
This conservative calibration acts as a \textbf{high-pass filter} for grounding. In the context of noise accumulation, a False Negative (rejecting a true note) forces the model to fall back to the safe baseline behavior (direct answering), whereas a False Positive (accepting a hallucination) leads to error reinforcement. Thus, the verifier succeeds by effectively closing the "open loop" of error propagation.

\textbf{Protocol Details.}
The verifier outputs one label from \texttt{SUPPORTED} / \texttt{UNSUPPORTED} / \texttt{UNCERTAIN}.
We apply a deterministic gate: $\text{keep}(n)=\mathbb{1}[\texttt{verifier}(I,q,n)=\texttt{SUPPORTED}]$.
This choice intentionally biases toward precision.
As shown in Table~\ref{tab:chartqapro_verified_control}, this minimal verification largely recovers the naive single-board drop across raw, answerable-only, and post-processed metrics.

\paragraph{Scope of Verification (Why ChartQAPro?).}
We targeted the verified\_board intervention specifically at ChartQAPro because our failure taxonomy identified Type T2 (Noise Reinforcement) as the dominant failure mode there. Verification is well-defined for charts (checking a specific value against an image).
In contrast:
\begin{itemize}
    \item VQAonline: The primary failure is Type T4 (Policy Collapse), which is a deficiency in output style/coverage rather than factual correctness. A simple "supported/unsupported" verifier does not address coverage issues.
    \item SlideVQA: This is a high-recall retrieval task. Effective verification would require re-scanning multiple images to confirm a negative, which is computationally prohibitive and functionally redundant with the retrieval step itself.
\end{itemize}
Thus, while verification is a powerful tool for arresting hallucination (ChartQAPro), different mechanisms (e.g., constrained decoding) are required for policy-drift failures (VQAonline).

\section{Two-stage Baseline (Diagnostic)}

\subsection{Two-stage evidence$\rightarrow$answer baseline}
\label{sec:two_stage_prompt}

This appendix specifies the two-stage single-agent baseline reported in Table~2 (marked $^{\ddagger}$).
The goal is to isolate the effect of multi-step test-time prompting from shared-board collaboration: the baseline uses no board and no inter-agent communication, but introduces an additional evidence-oriented call before the final answer.
All evaluation uses the same output parsing and no-exclusion scoring policy as the main runs.

\paragraph{Protocol.}
Given an instance $(I, q)$, the two-stage baseline makes exactly two model calls:
(1) an \textbf{evidence stage} that extracts a compact, checkable evidence summary; and
(2) a \textbf{final-answer stage} that produces the scored prediction conditioned on $(I, q)$ and the evidence summary.

\paragraph{Backbone and decoding.}
We use the same backbone as the corresponding main setting. Both stages use deterministic decoding (temperature $=0$) and the same per-call decoding cap (max\_new\_tokens$=32$).

\paragraph{Prompts.}

\noindent\textbf{Stage 1: Evidence extraction prompt (no answer).}
\begin{verbatim}
Task: extract evidence only. 
Do NOT answer the question.
Output format (strict):
EVIDENCE:
- <bullet 1>
- <bullet 2>
\end{verbatim}

\noindent\textbf{Stage 2: Final answer prompt (answer only).}
\begin{verbatim}
EVIDENCE SUMMARY: {EVIDENCE_FROM_STAGE_1}
Answer the question using ONLY the evidence 
summary and what is visible in the image.
Output format (strict):
FINAL: <answer>
\end{verbatim}

\section{Additional Qualitative Evidence}
\label{sec:appendix_qualitative}

\subsection{Case Studies}
\label{sec:appendix_case_studies}

\begin{figure}[t]
\centering
\small
\setlength{\fboxsep}{6pt}
\setlength{\fboxrule}{0.8pt}

\noindent
\begin{minipage}[t]{0.48\linewidth}
  \vspace{0pt} 
  \centering
  \raisebox{-\height}{%
    \includegraphics[width=\linewidth, trim=0 420 0 120, clip]{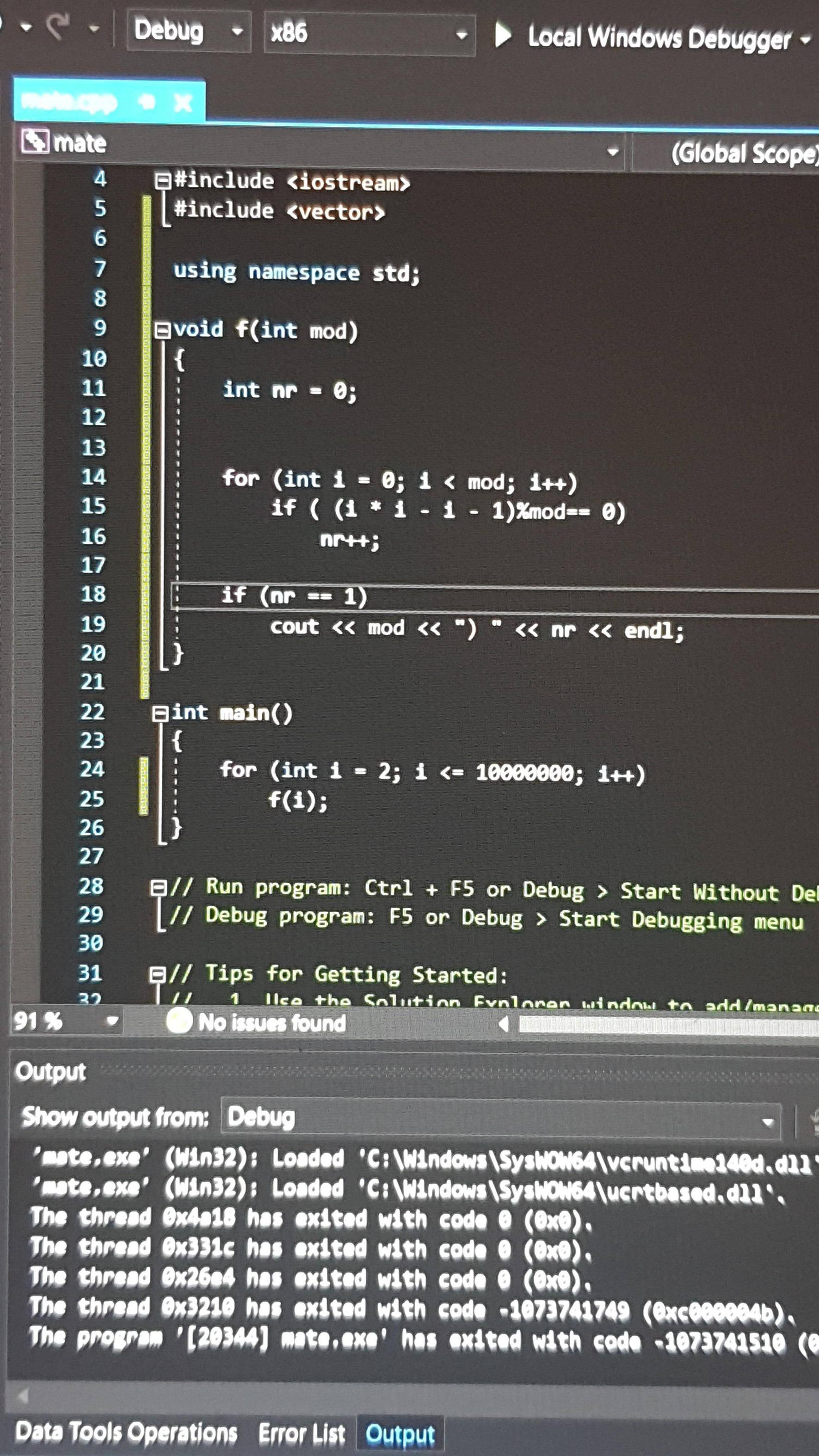}%
  }
\end{minipage}\hfill
\begin{minipage}[t]{0.48\linewidth}
  \vspace{0pt} 
  \fbox{%
    \parbox[t]{\dimexpr\linewidth-2\fboxsep-2\fboxrule}{%
      \raggedright
      \textbf{Case A: VQAonline (T4)}\\
      \textit{Failure: Output-policy drift}\\[4pt]
      \textbf{\textsc{Query}:} Solutions of $x^2=x+1$ in $\mathbb{Z}/n\mathbb{Z}$.\\
      \textbf{\textsc{Gold}:} Long-form derivation/characterization.\\
      \textbf{\textsc{Pred}:} \texttt{B} (near-empty).\\
      \vspace{2pt}
      \hrule width 0.5\linewidth height 0.4pt \vspace{2pt}
      \textbf{\textsc{Mechanism}:} Intermediate notes do not translate into final coverage.
    }%
  }
\end{minipage}

\vspace{4mm}

\noindent
\begin{minipage}[t]{0.48\linewidth}
  \vspace{0pt}
  \centering
  \raisebox{-\height}{%
    \includegraphics[width=\linewidth, trim=0 520 0 220, clip]{figures/chart_000069.png}%
  }
\end{minipage}\hfill
\begin{minipage}[t]{0.48\linewidth}
  \vspace{0pt}
  \fbox{%
    \parbox[t]{\dimexpr\linewidth-2\fboxsep-2\fboxrule}{%
      \raggedright
      \textbf{Case B: ChartQAPro (T2)}\\
      \textit{Failure: Mis-grounded intermediate}\\[4pt]
      \textbf{\textsc{Query}:} \emph{Approx.\ \# people who strongly favor the death penalty (round down).}\\
      \textbf{\textsc{Gold}:} 1379 \quad \textbf{\textsc{Pred}:} 27.\\
      \textbf{\textsc{Board}:} ``27\% strongly favor'' (percentage).\\
      \vspace{2pt}
      \hrule width 0.5\linewidth height 0.4pt \vspace{2pt}
      \textbf{\textsc{Mechanism}:} The board captures the wrong variable (rate vs.\ count), and the final answer inherits it.
    }%
  }
\end{minipage}

\caption{\textbf{Qualitative analysis of dominant failure mechanisms.}
\textbf{Top (Case A):} On VQAonline, the presence of board context induces a policy collapse, shifting the model toward overly terse responses (Type T4).
\textbf{Bottom (Case B):} On ChartQAPro, the model hallucinates a unit (treating percentage as count) and writes it to the board, reinforcing the error in the final step (Type T2).}
\label{fig:case_studies}
\end{figure}

\paragraph{Case studies.}
Figure~\ref{fig:case_studies} provides representative examples for the two dominant failure mechanisms surfaced by our taxonomy.
On VQAonline, the \texttt{single\_board} setting often collapses to overly terse final outputs (T4), consistent with the strong coupling between answer length and token-F1.
On ChartQAPro, intermediate notes can be mis-grounded (T2): the shared board records a salient statistic but misinterprets the variable (e.g., rate vs. count), hardening the error.

\section{Failure Taxonomy}
\label{sec:taxonomy_protocol}

\subsection{Sampling and annotation protocol}
For each dataset and method, we construct an error pool from the fixed evaluation subset.
From each pool, we uniformly sample $100$ error examples \emph{without replacement} to form the taxonomy set.
Each case includes the input image, question, final prediction, and full trace artifacts.

\paragraph{Labeling procedure.}
Each case is assigned exactly one dominant failure type from T0--T6 using a deterministic rubric, prioritizing the \emph{earliest causal mechanism}:
(1) parsing/format failure (T0);
(2) artifact leakage (T1);
(3) board-amplified wrong note (T2);
(4) attention dilution (T3);
(5) output-policy drift/under-coverage (T4);
(6) decision failure (T5);
(7) residual perception error (T6).

\subsection{Failure type definitions}
\label{sec:taxonomy_definitions}

\paragraph{T0: Missing or invalid final answer.}
The pipeline fails to extract a valid final prediction string (e.g., empty string). Scored as incorrect.

\paragraph{T1: Note/trace leakage.}
The final answer is dominated by intermediate artifacts (e.g., \texttt{Board:}, \texttt{Note:}) rather than an actual response.

\paragraph{T2: Board-amplified wrong intermediate (Noise Reinforcement).}
A wrong intermediate note is written to the board and ``hardened'' into the final answer. Example: Chart shows 50 (count), Note says "50\%", Final Answer says "50\%".

\paragraph{T3: Attention dilution.}
The trace shows agents attending to irrelevant evidence or conflicting notes without convergence, resulting in inconsistent reasoning.

\paragraph{T4: Output-policy drift (Policy Collapse).}
The final answer is materially under-specified relative to the question, driven by a shift toward overly short responses (common on VQAonline).

\paragraph{T5: Decision failure.}
Mishandling of answerability (e.g., answering an unanswerable question or vice versa).

\paragraph{T6: Residual perception error.}
Standard OCR or grounding failures not specific to the protocol (no note leakage or policy drift).

\section{Implementation Details and Prompts}
\label{sec:implementation}

\subsection{Reproducibility Statement}
\label{sec:reproducibility}
\textbf{(Response to reviewer inquiry on replication)}
To enable full third-party replication and auditing of weak-learner dynamics, we will release the following artifacts upon publication:
\begin{itemize}
    \item \textbf{Codebase:} Full PyTorch/HuggingFace implementation of the \textsc{CoSee} controller, agent roles, and evaluation pipeline.
    \item \textbf{Trace Logs:} The complete JSONL logs for all main experiments (approx. 200MB), including every board state, intermediate note, and audit flag. This allows future work to analyze error dynamics without re-running inference.
    \item \textbf{Verified Exemplars:} The specific subset of board notes accepted vs.\ rejected by the verifier, serving as a dataset for studying hallucination detection in charts.
    \item \textbf{Subset IDs:} The exact example IDs used for the fixed 600/800-sample evaluation subsets.
\end{itemize}

\subsection{Prompt Templates}
We use a unified prompt structure across all backbones (Qwen3, Phi-4, Gemma-3). The system instruction is prepended to the user query.

\paragraph{Board-Writing Prompt (Scanner / Single-Agent).}
Used for generating intermediate notes.
\begin{verbatim}
You are a helpful assistant.
IMAGE: [Image Features]
BOARD STATE: [Previous Notes (if any)]
QUESTION: {question}
INSTRUCTION: Identify specific visual 
evidence relevant to the question.
Write a concise note. Do not answer 
the question yet.
FORMAT: "Note: <content>"
\end{verbatim}

\paragraph{Verification Prompt (Verified-Board).}
Used for the S6.2 ChartQAPro control.
\begin{verbatim}
Task: Verify if the claim is directly 
supported by the image.
CLAIM: {note_content}
IMAGE: [Image Features]
QUESTION: {question}
INSTRUCTION: Look at the image.
Is the claim supported by visible 
evidence?
Output strictly one token: SUPPORTED, 
UNSUPPORTED, or UNCERTAIN.
\end{verbatim}

\paragraph{Final Answer Prompt.}
Used for all protocols (Single-Board, CoSee, Baseline) to generate the final prediction.
\begin{verbatim}
You are a helpful assistant.
IMAGE: [Image Features]
BOARD HISTORY:
{board_summary}
QUESTION: {question}
INSTRUCTION: Answer the question based 
on the image and the board history.
If the board contains errors, prioritize 
the image.
FORMAT: "Final Answer: <answer>"
\end{verbatim}

\subsection{Analysis of Grounding Pointers}
\label{sec:pointer_analysis}
\textbf{(Response to reviewer inquiry on pointer usage)}
A structural limitation of the weak learners evaluated (4B--8B parameters) is their struggle to produce precise visual coordinates (bounding boxes) even when explicitly prompted. In our qualitative audit of the \texttt{single\_board} traces on ChartQAPro:
\begin{itemize}
    \item \textbf{Pointer Scarcity:} Valid grounding pointers (e.g., box coordinates matching a chart element) appeared in less than $5\%$ of intermediate notes.
    \item \textbf{Hallucinated Precision:} When forced to output coordinates (via a "pointer-required" ablation prompt), the models frequently generated boxes covering the entire image or irrelevant regions, which did not mitigate T2 errors.
\end{itemize}
This observation explains the high prevalence of \textbf{Type T2 (Board-Amplified Error)} failures: without reliable geometric grounding, the board relies solely on textual semantic matching, which is prone to hallucination. This negative result motivates our specific design of the \texttt{verified\_board} control (Section~\ref{sec:verified_board}), which uses an image-to-text verification step as a proxy for missing geometric grounding.

\end{document}